\documentclass[10pt,twocolumn,letterpaper]{article}

\usepackage{iccv}
\usepackage{times}
\usepackage{epsfig}
\usepackage{graphicx}
\usepackage{amsmath}
\usepackage{amssymb}
\usepackage{booktabs}
\usepackage{url}
\usepackage{graphicx}
\usepackage{epstopdf}
\usepackage{cite}
\usepackage{gensymb}
\usepackage{url}
\usepackage[bottom]{footmisc}
\usepackage{enumitem}
\usepackage{multirow} 
\usepackage{color}
\usepackage{subfig}
\usepackage{breqn}
\usepackage{caption}
\usepackage{tabularx}
\usepackage{fixltx2e}
\usepackage[english]{babel}
\def\Vec#1{{\boldsymbol{#1}}}

\renewcommand{\Vec}[1]{\boldsymbol{\mathbf{#1}}}

\usepackage[pagebackref=true,breaklinks=true,letterpaper=true,colorlinks,bookmarks=false]{hyperref}

 \iccvfinalcopy 

\def\iccvPaperID{1875} 

\ificcvfinal\fi
\begin{document}

\title{A Convolution Tree with Deconvolution Branches: Exploiting Geometric Relationships for Single Shot Keypoint Detection}

\author{  Amit Kumar \hspace{15pt} Rama Chellappa\\
Department of Electrical and Computer Engineering, CFAR and UMIACS\\
     University of Maryland-College Park,USA \\
{\tt\small akumar14@umiacs.umd.edu, rama@umiacs.umd.edu}
}

\maketitle

\begin{abstract}
Recently, Deep Convolution Networks (DCNNs) have been applied to the task of face alignment and have shown potential for learning improved feature representations. Although deeper layers can capture abstract concepts like pose, it is difficult to capture the geometric relationships among the keypoints in DCNNs. In this paper, we propose a novel convolution-deconvolution network for facial keypoint detection. Our model predicts the 2D locations of the keypoints and their individual visibility along with 3D head pose, while exploiting the spatial relationships among different keypoints. Different from existing approaches of modeling these relationships, we propose learnable transform functions which captures the relationships between keypoints at feature level. However, due to extensive variations in pose, not all of these relationships act at once, and hence we propose, a pose-based routing function which implicitly models the active relationships. Both transform functions and the routing function are implemented through convolutions in a multi-task framework. Our approach presents a single-shot keypoint detection method, making it different from many existing cascade regression-based methods. We also show that learning these relationships significantly improve the accuracy of keypoint detections for in-the-wild face images from challenging datasets such as AFW and AFLW. 
\end{abstract} 
\section{Introduction}
A key step towards tasks such as face modelling, verification and recognition, driver monitoring in autonomous driving systems is accurate keypoint estimation of human face image. Given an RGB face image we wish to determine the precise locations of important keypoints. In the past few years many methods have been proposed to address this problem mostly taking the cascade regression approach. While, earlier methods such as\cite{DBLP:journals/ijcv/CaoWWS14},\cite{DBLP:conf/cvpr/RenCW014},\cite{akshay_wild},\cite{6909635},\cite{kazemi2014one} were dependent on hand-crafted features, recently methods based on DCNNs\cite{Zhu_2015_CVPR},\cite{DBLP:conf/eccv/ZhangLLT14},\cite{DBLP:journals/corr/KumarRPC16},\cite{2017arXiv170205085K} have become very prominent. Despite a long history of research on keypoint extraction, the problem of keypoint detection for in-the-wild faces still remains an open challenge, mainly due to extreme variations in pose, image quality and unavailability of large scale annotated in-the-wild datasets. 
\begin{figure}[t]
 \centering
\includegraphics[width=8cm,height=2cm]{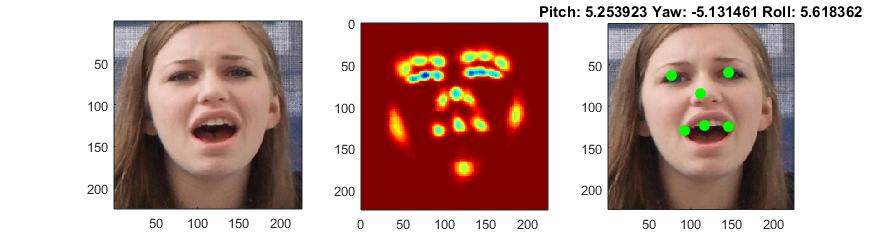}
\caption{Sample result generated using the proposed method on an image from AFW dataset. The second picture is the generated response map. Third picture shows the predicted points including the invisible points, indicating the effectiveness of message passing between different fiducial points. The predicted 3D pose is also mentioned.}
\label{fig:illustration}
\end{figure}

On a related note, a similar task is estimating the joint locations for human pose estimation, where an approach taken by most of the researchers is based on directly regressing over the heatmaps predicted by DCNNs. These methods have shown tremendous potential in capturing feature representations by achieving state-of-the art performances in human pose estimation problems\cite{7780704}. The heatmaps provide a probability value, indicating the existence of a certain joint at a specific location. \cite{5995741},\cite{Chen_NIPS14} have shown that by modeling the geometric relationships, independent predictions of the keypoints can be refined. While these methods have shown to be effective for the prediction of human body joints, predicting facial keypoints (also called fiducials) presents even greater challenge due to extensive variations in pose, expression, individual visibility of keypoints, lightning, occlusion and identity. Due to this, there have been fewer attempts\cite{recdec} at addressing the facial keypoint estimation problem by regressing over the heatmaps. 

Modeling the geometric relationships between the fiducials presents a great challenge in a deep learning-based framework as the invisible points are not annotated in profile face images. Human perception works in a way that one can tell the location of a specific keypoint conditioned on the location of other points. Even for an invisble keypoint, one can infer with confidence the likely location of the keypoint. This comes naturally as a result of the perceived head pose and inherent geometric relationship among the keypoints. \cite{chu2016structure} has shown that if all the keypoints are annotated such a geometric relationship can be modeled by some learned transform kernels. For this \cite{chu2016structure} proposed a bi-directional tree with convolutions. However in \cite{chu2016structure} it was assumed that all the keypoints communicate with each other irrespective of their visibilities in the image, which may not be true for in-the-wild face images. Therefore, to capture the relationship that exists between 3D pose and the visibility of individual keypoint, we formulate the keypoint detection problem conditioned on head pose. Along with this, we use convolutions to capture the spatial relationships between pairs of keypoints. Figure \ref{fig:illustration} shows a sample result generated from the proposed method.  

In this paper, we introduce a new deep network composed of a convolution network with deconvolution branches trained in a multi-task framework with pose-dependent routing function which implicitly controls the message passed through invisible keypoints. For message passing between different keypoints, we propose a tree structure in which keypoints receive information from neighboring points. Depending on the head pose, the routing function which acts on deeper layers of the convolution tree implicitly activates only certain branches of the deconvolution layers. This is possible as convnets employ multiple layers to learn hierarchical feature representations of input images. Features in lower layers capture low-level information, while those in higher layers can represent more abstract concepts such as pose and attributes. Both message passing and the routing functions are implemented by non-linear transformations of convolution outputs. To refine the keypoints locations and their visibilities, regression networks are concatenated and trained in an end to end fashion. In addition we use the Residual Squeezenet\cite{SqueezeNet}, which greatly reduces the number of parameters while retaining the advantages of a residual network. Our proposed method is non-iterative and fully convolutional and hence, is fast.  

We show the efficiency of our proposed method on two public and challenging datasets widely used for face alignment benchmarks: AFW\cite{AFW_dataset_CVPR2012} and AFLW\cite{tugraz:icg:lrs:koestinger11b}. The proposed method is a general framework and can also be trained for other tasks. The main contributions of this paper are three folds:
\begin{itemize}
\item We design a novel end to end learning framework with convolution and deconvolution layers which effectively captures the message passing between different fiducial points. 
\item The message passing between fiducial points is conditioned on head pose by a learnable routing function. 
\item We include a regression network to obtain the final location of keypoints. Along with locations of keypoints our network also predicts the 3D head pose and individual visibility of each fiducial point. Exploiting the benefits of residual connections and fully convolutional networks, the methods predicts the keypoint locations in a single pass and hence is faster than other cascade deep regression methods. 
\end{itemize}

\section{Related work}
Face alignment has advanced in the last five years after the reemergence of deep neural networks. Following \cite{DBLP:journals/ijcv/CaoWWS14}, we classify previous works on face alignment into two basic categories. 

\textit{Part-Based Deformable} model approaches are statistical methods which perform keypoint detection by maximizing the confidence of part locations in a given input image. Zhu and Ramanan\cite{AFW_dataset_CVPR2012} used a part-based model for face detection, pose estimation and landmark localization assuming the face shape to be a tree structure.\cite{Asthana:2013:RDR:2514950.2516059} by Asthana et al. proposed learning a dictionary of probability response maps followed by linear regression in a Constrained Local Model (CLM) framework. Other works in this category include Active Shape Models\cite{Cootes:1995:ASM:206543.206547} and Constrained Local Models\cite{Cristinacce:2008:AFL:1385702.1385969}. To enforce the message passing protocol between different keypoints the proposed method assumes a tree structure of the keypoints; however does not assume that all the keypoints are visible and contributing equally to each other. In the proposed tree structure the messages between neighboring keypoints are passed via learnt transform kernels which are further conditioned on the 3D head pose of the face image. \par
\begin{figure*}[ht!]
\centering
\includegraphics[height=6cm,width=15cm]{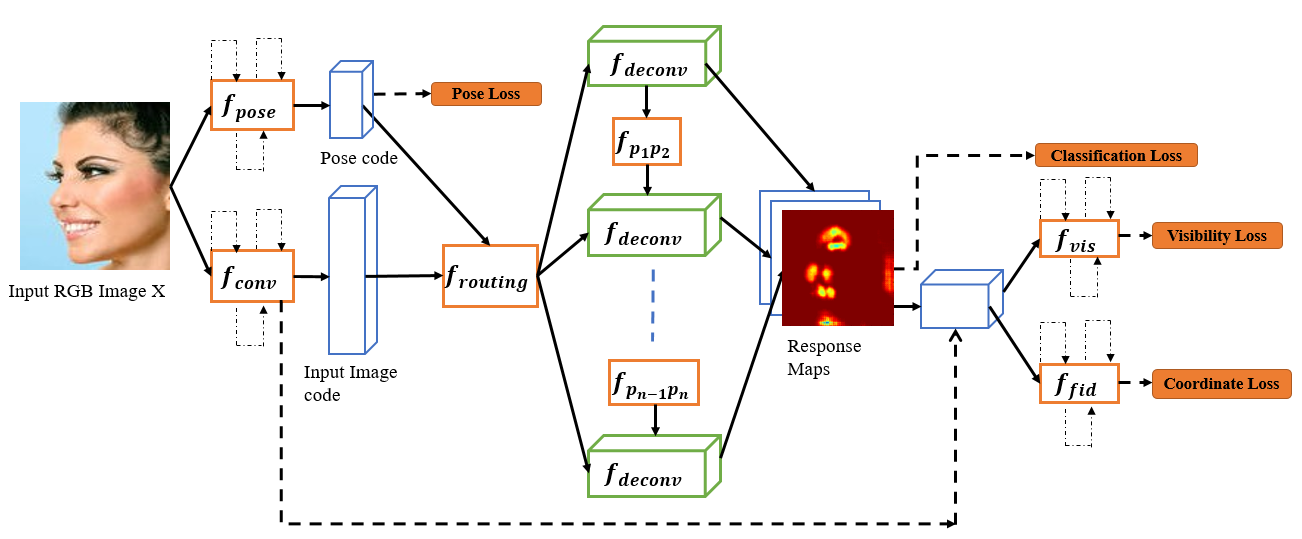}
\caption{Overview of the proposed method. The dotted lines on top of $f_{pose}$, $f_{conv}$, $f_{vis}$ and $f_{fid}$ denotes residual connections. The dotted line between $f_{conv}$ and $f_{reg}$ indicates feature concatenation. $f_{pose}$, $f_{conv}$, $f_{deconv}$, $f_{p_{i}p_{j}}$, $f_{vis}$ and $f_{fid}$ are non-linear multi-layered mappings.}
\label{fig:outline}
\end{figure*}

In a \textit{Cascade Regression-based} framework, image appearances are directly mapped to output space producing keypoint coordinates. Recently, a multitude of cascade regression-based methods\cite{DBLP:journals/ijcv/CaoWWS14},\cite{Zhu_2015_CVPR} significantly boost the keypoint detection performance, compared to statistical methods described above. However, these methods along with methods from \cite{trigeorgis2016mnemonic},\cite{akshay_wild},\cite{6909635} and \cite{6619290} were mostly evaluated on face images where all the facial keypoints are visible. To handle occlusion, Wu et al.\cite{7410774} proposed an  occlusion-robust cascaded regressor. Supervised Descent Method (SDM)\cite{XiongD13} learns a cascade of regression models based on SIFT features. To mitigate the conflicting gradient directions in SDM, Xiong et al.\cite{global2015xiong} suggested domain dependent descent maps. Inspired by\cite{global2015xiong}, Cascade Compositional Learning (CCL)\cite{Zhu_2016_CVPR} developed a head pose based method by partitioning the optimization domain. Different from all these methods, our approach is a non-iterative single shot method, which along with keypoint locations also provides the estimated 3D head pose and individual visibility of each fiducial point.

Researchers also proposed using 3D morphable models to estimate the landmark points. Pose Invariant Face Alignment (PIFA)\cite{pifa} by Jourabloo et al. suggested a 3D model-based approach that employed cascaded regressors to predict the coefficients of 3D to 2D projection matrix. \cite{lfa3d} also by Jourabloo et al. formulated the face alignment problem as a dense 3D model fitting problem, where the camera projection matrix and 3D shape parameters were estimated by a cascade of CNN-based regressors. However, \cite{Zhu_2016_CVPR} suggests that optimizing  the  base  shape coefficients and projection is indirect and sub-optimal since smaller parameter errors are not necessarily equivalent to smaller alignment errors. 3DDFA\cite{DBLP:journals/corr/ZhuLLSL15} modeled depth data in Z-Buffer and fitted a dense 3D face model to the image via CNNs. In contrast to these methods the proposed method does not rely on 3D morphable models, but still provides with accurate 3D pose estimates.

The proposed method uses both a classification and regression framework in a non-iterative way and hence does not directly fall into any of the above categories. To estimate the intermediate response map, we use a classification loss which is further regularized by regression loss. All the four tasks are trained end-to-end in a multitask framework. We use regression for head pose estimation which further modulates the performance of the routing function. One of the closely related work is \cite{recdec}, which also uses classification and regression together to estimate the keypoint locations. However, \cite{recdec} does not take into consideration the spatial relationships between keypoints and hence goes for an iterative solution.
\section{Convolution Tree with Deconvolution Branches}
In this section we first give an overview of the proposed approach. Then we describe constrained joint training in multi-task framework. 
\subsection{Method Overview}
The task of keypoint detection is to estimate the 2D coordinates of, say \textit{L} landmark points, given a face image. Observing the effectiveness of deep networks for variety of vision tasks, we formulate our task in an end-to-end trainable deep neural network.
Figure \ref{fig:outline} shows the overview of our approach. In the rest of the paper, we consider \textit{f\textsubscript{*}} as a multi-layered non-linear function. Unlike KEPLER\cite{2017arXiv170205085K}, the input to our network is RGB image \textbf{x}$\in\mathbb{R}^{w\mathrm{x}h\mathrm{x}3}$ while the label map \textbf{z}$\in\mathbb{R}^{w\mathrm{x}h\mathrm{x}(L+1)}$. Each $1\mathrm{x}1\mathrm{x}(L+1)$ block in the label map \textbf{z} is vector label $\{0,\ldots,L+1\}$ where $L$ is the number of keypoints and one extra channel added for background. 

The function \textit{f\textsubscript{conv}} performs a series of convolutions, pooling operations, residual additions and non-linear transformations to extract a lower-dimensional feature representation of the image. The features thus obtained are referred to as the image code $\textbf{C}_{\textbf{x}}$ in figure \ref{fig:outline}, while the parameters of this network are denoted by $\theta_{conv}$:
\begin{equation}
\textbf{C}_{\textbf{x}} = f_{conv}(\textbf{x},\theta_{conv})
\label{eq1}
\end{equation}
Similarly, the the function \textit{f\textsubscript{deconv}} performs a series of strided convolutions, non-linear operations and batch normalization to produce a response map of each fiducial point. 
\begin{equation}
\textbf{R}_{\textit{i}} = f_{deconv}(C_{\textbf{x}},\theta_{deconv_{i}},\theta_{m_{i,j}})
\label{eq2}
\end{equation}
where $\textbf{R}_{\textit{i}}$ and $\theta_{deconv_{i}}$ refers to the the response map and deconvolution parameters, respectively of $\textit{i}^{th}$ fiducial point. $\theta_{m_{i,j}}$ are the parameters of the message passing layer between the $\textit{i}^{th}$ and $\textit{j}^{th}$ fiducial point. 
\begin{figure}[htp]
\centering
\includegraphics[width=8cm,height=4cm]{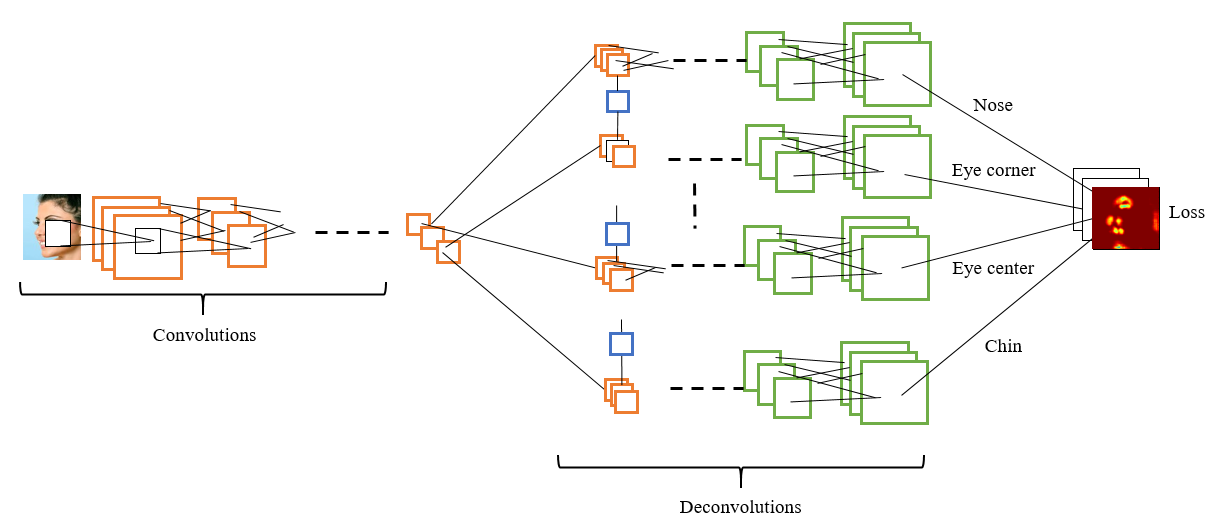}
\caption{A simplified version of the convolution tree with deconvolution branches.}
\label{fig:network}
\end{figure}

The convolution tree with deconvolution branches plays an important role in the task of fiducial point detection. First, it forms a low-dimensional embedding of the input image and generates an output of the same size as the image but with different number of channels. This enables us to put the regression module on top of the deconvolution outputs for finer localization. Next, the lower dimensional embedding of the input image, enables the pose prior module to implicitly control the flow of message between different deconvolution branches. Since the output maps after deconvolution are of the same size as of the image, there is no loss of information in upsampling of the output response maps.  
\subsubsection{Message Passing and Routing Function}
\begin{figure}[htp]
\centering
\includegraphics[width= 7.5cm,height =3.5cm]{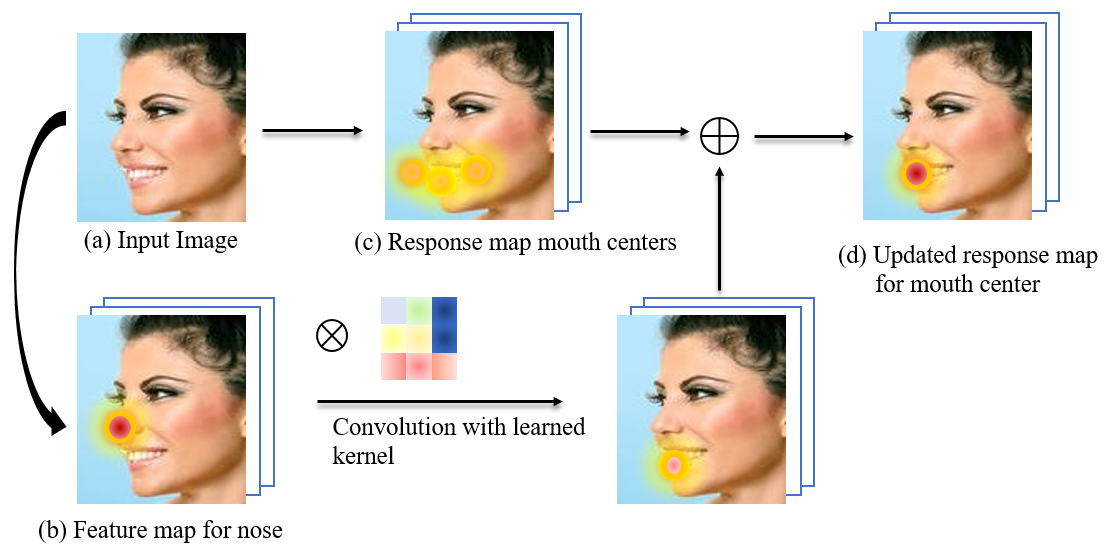}
\caption{Illustration showing how response maps can be updated by passing information between keypoints.}
\label{fig:message}
\end{figure}
Spatial distributions and semantic meaning of feature maps obtained for different fiducial points are correlated, and hence information from one keypoint can be passed to other neighboring points and effectively improve quality of the feature representation at each keypoint. Inspired by\cite{chu2016structure}, we proceed in the same direction and implement the message passing protocols by convolutions. We illustrate the message passing process in Figure \ref{fig:message}. Given an input image (a), the feature map for `nose tip' is shown in (b). Also three candidate response maps for `mouth centers' are shown in (c). In (d), we show the updated confidence of the location of `mouth center' after addition of candidate `mouth center' responses with the convolved response map of `nose tip' with a learned transform function $f_{p_{i}p_{j}}$.  

\begin{figure}[htp]
\centering
\includegraphics[width= 4cm,height =4cm]{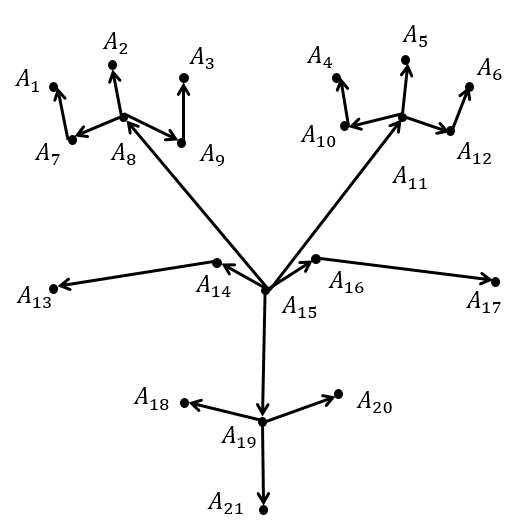}
\caption{The proposed tree structure for effective information passing between fiducial points.}
\label{fig:tree}
\end{figure}

One expects to receive information from all other keypoints in order to optimize the features at a specific keypoint. However, this has two drawbacks: First, to model the information of keypoints far away such as `eye corner' and `chin', transform functions with larger size have to be introduced. This also leads to increase in the number of parameters. Secondly, relationships between some keypoints are unstable, such as `left eye corner' and `right eye corner'. In a profile face image one of the points may not be visible and passing information between these two fiducial points may lead to erroneous results. Hence, we introduce the tree structure for the keypoints, shown in figure \ref{fig:tree}. Keypoints which are closer and have stable relationships are connected together Figure \ref{fig:tree} also shows the flow of information assuming `nose tip' to be the root node.  

Such a tree structure was also used in \cite{chu2016structure} to model the skeletal joints for the task of human pose estimation. However, \cite{chu2016structure} did not take into account the fact that all of the points may not be influencing other points at the same time for example, in case of occlusion. This is even more prominent for face images which are captured in challenging conditions. This flow of information is highly dependent on the 3D pose of the face. Due to the hierarchical nature of deep convolution networks, the deeper layers possess more abstract information such as pose, attributes and object categories\cite{DBLP:journals/corr/ZeilerF13}. Therefore, in parallel, we also perform a sequence of convolution, non-linear, pooling and batch normalization operations on the RGB image to predict the 3D head pose. Pooling information from deeper layers of this parallel network and adding it to the feature representation from the convolution tree, is how the network tries to implicitly control the information flow through the deconvolution branches.     
\subsubsection{Constrained Shape Prediction}
In cases where distractions exist, such as another partial face inside the bounding box, the response heatmaps may have few false positives. Although this issue is significantly reduced by the message passing layers which impose shape constraint explicitly, it still degrades the performance in challenging situations. Hence, we append a constrained shape regression network which takes the response map as input and finally predicts the precise keypoint locations. This regression network jointly predicts the coordinates of all the points while maintaining the shape constraint as in\cite{DBLP:journals/ijcv/CaoWWS14},\cite{2017arXiv170205085K}. 

\textit{f\textsubscript{fid}} outputs the keypoint coordinates $\Vec{y}\in\mathbb{R}^{2L\mathrm{x}1}$, where the gradients are not back propagated for the invisible keypoints. All the coordinates are normalized by subtracting a mean shape obtained from the training images. 

\subsubsection{Loss Functions}
\label{loss}
The loss function for each of the tasks is discussed below. \par
\textbf{Classification loss:}
In order to train the network, the localization of fiducial keypoints is formulated as a classification problem. The label for an input image of size $h\mathrm{x}w\mathrm{x}3$ is a label tensor of same size as the image with $L+1$ channels, \textit{i.e.} $224\mathrm{x}224\mathrm{x}22$ in our case. The first $21$ channels represents the location of each keypoint whereas the $22^{nd}$ channel represents background. Each pixel is assigned a class label, where the objective is to minimize the following loss function:
\begin{equation}
L_{0}(\Vec{p},\Vec{g}) = \sum_{i=1}^{h}\sum_{j=1}^{w}m(i,j)\sum_{k=1}^{L+1}g_{k}(i,j)log(\frac{e^{p_{k}(i,j)}}{\sum_{l}e^{p_{l}(i,j)}})
\label{eq3}
\end{equation} 
where $k\in[1,2\ldots 22]$ is the class index and $g_{k}(i,j)$ represents the ground truth at location $(i,j)$. $p_{l}(i,j)$ is the score obtained for location $(i,j)$ after forward pass through the network. Since the number of negative example is much larger than the positives, we design a mask $m(i,j)$ which keeps only $0.025\%$ of the negative samples by random selection. \par
\textbf{Coordinate Regression loss:}
For coordinate loss we use the square of differences loss between predicted and ground-truth locations. With each point is associated the visibility of that point. The loss function for this task is given by  
\begin{equation}
L_{1}(\Vec{y},\Vec{g}) = \sum_{i = 1}^{N} v^{i}(y^{i} - g^{i})^{2}
\label{eq4}
\end{equation} 
where $y^{i}$ and $g^{i}$ are the predicted and the ground truth locations of the $i^{th}$ keypoint respectively. $v^{i}$ is the ground truth visibility associated with each keypoint. This also ensures that there is no gradient back propagated for the invisible keypoints.\par
\textbf{Pose Prediction:}
Pose prediction refers to the task of estimating the 3D pose of the face. We use the Euclidean loss function for this task, where the loss is calculated as the sum of square differences between the predicted and ground-truth 3D pose. 
\begin{equation}
L_{2}(\Vec{p}_p,\Vec{g}_p) = (p_{yw}-g_{yw})^{2} + (p_{pch}-g_{pch})^{2} + (p_{rl}-g_{rl})^{2}
\label{eq5}
\end{equation} 
where $p$ stands for predicted and $g$ for the ground-truth 3D pose. $yw$, $pch$ and $rl$ indicates yaw, pitch and roll respectively. \par
\textbf{Visibility:}
This task is associated with estimating the visibility of each keypoint. The number of keypoints visible on the face varies with pose. Hence, we use the Euclidean loss to estimate the visibility confidence of each point.
\begin{equation}
L_{3}(\Vec{v}_p,\Vec{v}_g) = \sum_{i = 1}^{N} (v_{p,i} - v_{g,i})^{2},
\end{equation} 
where $v_{p,i}$ and $v_{g,i}$ denotes the visibility of $i^{th}$ keypoint. 

Therefore, the net loss in the network is the weighted linear combination of all the loss functions discussed above. 

\section{Network Architecture and Implementation Details}
All the modules are connected and trained in an end to end fashion. In this section we provide details about the network modules and the training procedure. 
\subsection{\textbf{\textit{f\textsubscript{{conv}}}} and \textbf{\textit{f\textsubscript{{deconv}}}}}
\begin{figure}[htp]
\centering
\includegraphics[width= 7.5cm,height = 1.8cm]{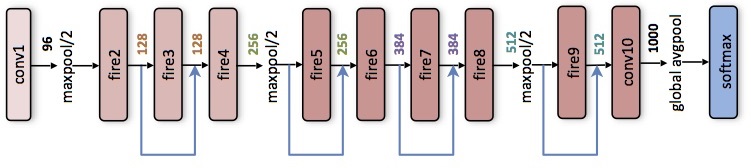}
\caption{The basic structure of residual squeezenet network composed of fire modules. }
\label{fig:squeezenet}

\end{figure}
\begin{figure}[htp]
\centering
\includegraphics[width= 8cm,height = 4.5cm]{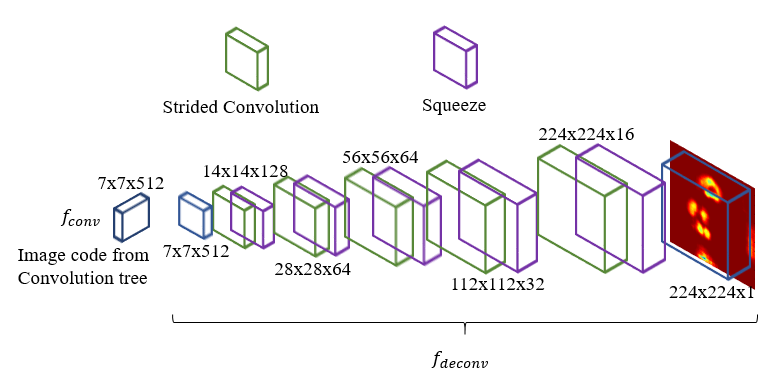}
\caption{Detailed description of a single deconvolution network. Each deconvolution network is identical to the one shown above.}
\label{fig:deconv}
\end{figure}
Figure \ref{fig:network} illustrates the \textit{f\textsubscript{conv}} and \textit{f\textsubscript{deconv}} schematically. The base structure of the convolution tree is designed as Squeezenet architecture as shown in figure \ref{fig:squeezenet}. The Squeezenet architecture consists of fire modules, where each fire module has squeeze and expand layers. By performing 1x1 convolutions, the squeeze layer first compresses the response from previous layer. Next, 1x1 and 3x3 convolutions with higher number of filters are performed to decompress the response from the squeeze layer. The expand response maps after 1x1 and 3x3 convolutions are then concatenated to form the final response of a fire module. This effectively reduces the number of parameters without getting hurt in performance. There are residual connections between some fire modules, where the output of previous layer is directly concatenated with the next layer. The choice of squeezenet in this work is based on the fact that the need of deconvolution branches increases the number of parameters $22$x. 

The deconvolution branches are also structured in a way similar to Squeezenet convolutions. In every upsampling operation, strided convolutions are performed. One convolution layer expands to a higher resolution while the next convolution squeezes the response using 1x1 convolution. Figure \ref{fig:deconv} illustrates the deconvolution operation, which upsamples the image using strided convolutions while maintaining fewer parameters using squeeze operations. 

To obtain the information from the $i^{th}$ keypoint, $3\mathrm{x}3$ convolutions were performed on the $7\mathrm{x}7$ response maps in the $i^{th}$ deconvolution branch. Non-linearity was then applied to the obtained output which was then added elementwise to the $7\mathrm{x}7$ response maps of $j^{th}$ branch. Experimentally it was found that $3\mathrm{x}3$ convolutions on $7\mathrm{x}7$ repsonse maps were sufficient to pass information effectively on the proposed tree structure. 
\subsection{\textbf{\textit{f\textsubscript{{routing}}}}}
\textit{f\textsubscript{{routing}}} is a composite function of the response from deeper layers of the pose network. A deeper layer, specifically pool8 of the squeezenet network is first convolved with 3x3 filters and then a non-linearity is applied, which is then added element-wise to the pool8 of the \textit{f\textsubscript{{conv}}}. It implicitly acts as a switch between the branches of the deconvolution networks through which information is passed. 
\subsection{\textbf{\textit{f\textsubscript{{pose}}}}, \textbf{\textit{f\textsubscript{{fid}}}} and \textbf{\textit{f\textsubscript{{vis}}}}}
\textit{f\textsubscript{pose}}, \textit{f\textsubscript{fid}} and \textit{f\textsubscript{vis}} are implemented by the Squeezenet network (shown in figure \ref{fig:squeezenet}) with their corresponding loss functions mentioned in section \ref{loss}. The dotted lines between \textit{f\textsubscript{conv}} and \textit{f\textsubscript{fid}},\textit{f\textsubscript{vis}} in figure \ref{fig:outline} represents concatenation of features from the convolution network to impart contexual features for improved regression.  

For initializing the weights of the network, we first pretrain the convolution tree with deconvolution branches for the classification task alone. This model is similar to the one shown in figure \ref{fig:network}. Starting from the learning rate of $1e^{-3}$, with momentum set to $0.9$ the baseline model was trained for $100K$ iterations. The multistep learning procedure was used, while the network was optimised with SGD solver. The convolution-deconvolution portion of the multitask network was then initialized with the weights obtained.
\section{Experiments and Results}
In this section we describe the datasets used for training and testing, evaluation protocols and evaluation metrics. 
\subsection{Datasets}
We select two challenging datasets with their most recent benchmarks.  

\textbf{\textit{In-the-wild datasets:}} We select AFLW\cite{tugraz:icg:lrs:koestinger11b} for training and, AFLW and AFW\cite{AFW_dataset_CVPR2012} as the main test sets. AFLW consists of face images with challenging shape variations, occlusion and significant view changes. This helps the system to perform robustly for images in real life scenarios. 

\textbf{AFLW} consists of $24,386$ in-the-wild faces (obtained from \textit{Flickr}) with head  pose ranging from $0\degree$ to $120\degree$ for yaw and upto $90\degree$ for  pitch and roll. AFLW also demonstrates significant external-object occlusion. There are a total of 21\% invisible landmarks caused by occlusion, larger than 13\% on COFW\cite{10.1109/ICCV.2013.191} where only internal object-occlusion is exhibited. 
In addition, COFW also provides the annotations for the invisible landmarks while in the case of AFLW the invisble landmarks are absent. 

\textbf{AFW} contains 468 in-the-wild faces ( also obtained from Flickr) with yaw degree up to $90\degree$ and is a popular benchmark for the evaluation of face alignment algorithms. The images are quite diverse in terms of pose, expression and illumination. The number of visible points also varies depending on the pose and occlusion. However, the locations of occluded points are to be predicted for images in the AFW testset.\\
AFLW provides at most 21 points for each face. It excludes coordinates for invisible landmarks and in our method such invisible points are labelled as background. In many cases such as in \cite{Zhu_2016_CVPR}, invisible points are hallucinated and re-annotated thereafter.\\ \\
\textbf{{Testing Protocols:}} \\ \\
\textbf{(I)AFLW-PIFA:} We follow the protocol used in PIFA\cite{pifa}. We randomly select $23,386$ images for training and the remaining $1,000$ for testing. We divide the test images in three groups with equal number of images in each group as done in \cite{pifa}: $[0\degree,30\degree]$, $[30\degree,60\degree]$ and $[60\degree,90\degree]$.\\ 
\textbf{(II)AFLW-Full:} We  also test on the full test set of AFLW of sample size $1,000$. \\ 
\textbf{(III)AFW:} We only use AFW for testing purposes. We follow the protocol as stated in \cite{AFW_dataset_CVPR2012}. AFW provides 468 images in total, out of which 329 faces have height and width greater than 150 pixels. We only evaluate on those 329 images following the protocol of \cite{AFW_dataset_CVPR2012}.\\ \\
\textbf{Evaluation metric:}  Following  most  previous  works, we obtain the error for each test sample via averaging normalized errors for all annotated landmarks. We illustrate our results with mean error over all samples, or via Cumulative Error Distribution (CED) curve. For pose, we evaluate on continuous pose predictions as well as their discretized versions rounded to nearest $15\degree$. We report the continuous mean absolute error for the AFLW testset and plot the Cumulative Error Distribution curve for AFW dataset. All the experiments including training and testing were performed using the Caffe\cite{jia2014caffe} framework and four Nvidia TITAN-X GPUs. Being a non-iterative and single shot keypoint prediction method, our method is fast and can process \textbf{7-8} frames per second on 1 GPU only. 
\begin{table}[htp]
\centering
\begin{tabular}{lll}
\hline
 & {{\textbf{\textit{AFLW}}}} & {{\textbf{\textit{AFW}}}} \\ 
\hline
{{\textbf{Method}}} & {{\textbf{NME}}}  & {{\textbf{NME}}}\\ \hline
TSPM \cite{AFW_dataset_CVPR2012}   & -     & 11.09   \\           
CDM \cite{Xiang_iccv_2013}         & 12.44    & 9.13          \\
RCPR \cite{10.1109/ICCV.2013.191}        & 7.85     &-          \\
ESR \cite{DBLP:journals/ijcv/CaoWWS14}         & 8.24    & -          \\
PIFA \cite{pifa}        & 6.8          & 9.42      \\
3DDFA \cite{DBLP:journals/corr/ZhuLLSL15}       & 5.32      &-         \\
LPFA-3D \cite{lfa3d}     & 4.72  & 7.43             \\
EMRT \cite{DBLP:journals/corr/ZhuLLT15}     & 4.01    & 3.55            \\ 
Hyperface \cite{DBLP:journals/corr/RanjanPC16}  & 4.26               \\
CCL \cite{Zhu_2016_CVPR}         & 5.85       & \textbf{2.45}        \\ 
Rec Enc-Dec\cite{recdec}  & \textgreater6        & -        \\ \hline 
\textbf{Ours}       & \textbf{3.93}   & 3.28    \\ \hline
\end{tabular}
\caption{Comparison of the proposed method with other state of the art methods. NME stands for normalized mean error. For AFLW, numbers for other methods are taken from respective papers following the PIFA protocol. For AFW, the numbers are taken from respective published works following the protocol of \cite{AFW_dataset_CVPR2012}.}
\label{aflw_table}
\end{table}

We would like to make a note that the testing protocols \textbf{I} and \textbf{II} are most relevant to this work since protocol \textbf{III} requires the prediction of invisible points. As mentioned in section \ref{loss} the pixel corresponding to the invisible points are labeled as background. This also enables the network to handle internal and external object occlusion. However, visual inspection of the results shows that ignoring the visibility, the prediction of invisible points from the heatmap is also very accurate. This shows the effectiveness of the explicit message passing layers between different keypoints.  
\begin{table}[h]
\centering
\resizebox{\columnwidth}{!}{%
\begin{tabular}{llll}
\hline
{\textbf{Method}} & {\textbf{AFLW-PIFA}} & {\textbf{AFLW-FULL}} & {\textbf{AFW}}\\ \hline
\textbf{Ours}   & 3.93 & 3.56 & 3.35     \\ \hline
\end{tabular}%
}
\caption{Summary of performance on different protocols of AFLW and AFW by the proposed method.}
\label{summary}
\end{table}
\begin{figure}[htp]
\centering
\includegraphics[height=5.5cm,width=8cm]{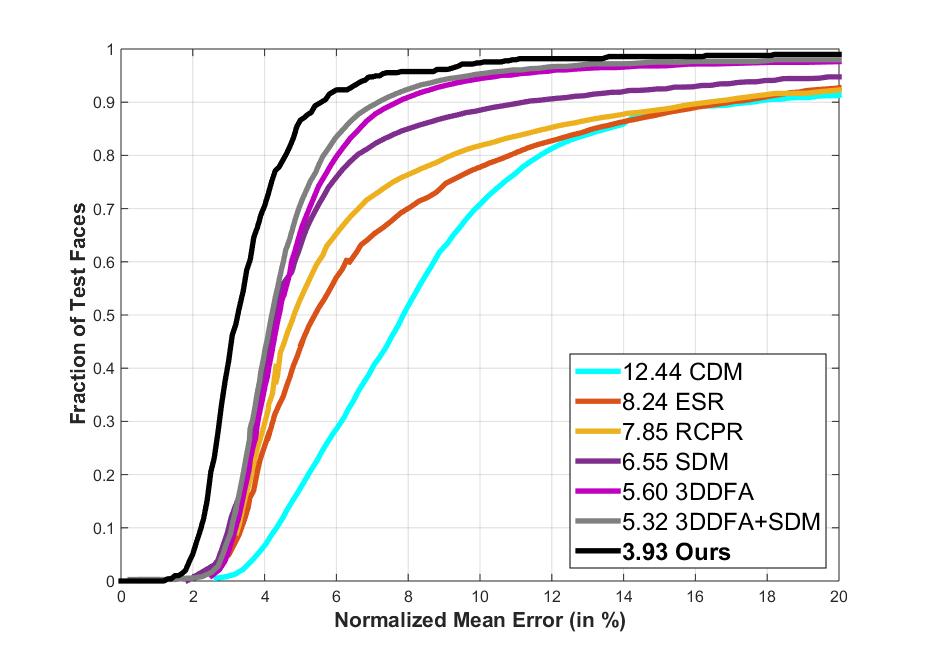}
\caption{Cumulative error distribution curves for landmark localization on
the AFLW dataset. The numbers in the legend are the average normalized mean error normalized by the face size.}
\label{aflw_res}
\end{figure}
\begin{figure}[htp!]
\centering
\includegraphics[height=5.5cm,width=8cm]{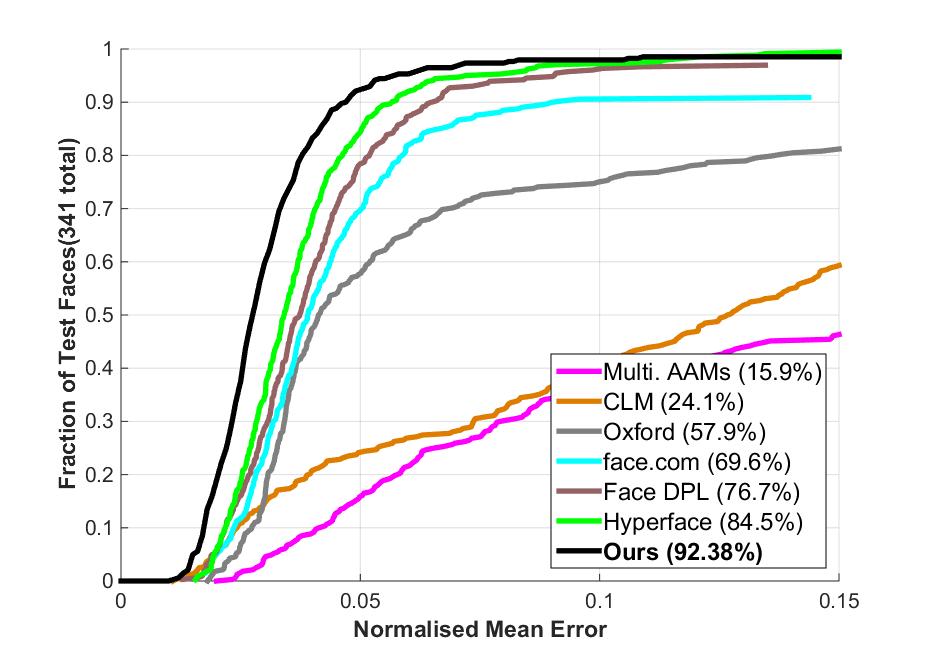}
\caption{Cumulative error distribution curves for landmark localization on
the AFW dataset. The numbers in the legend are the fraction of testing
faces that have average error below (5\%) of the face size.}
\label{afw_res}
\end{figure}
\begin{figure*}[t]
 \centering
\includegraphics[width=3cm,height=3cm]{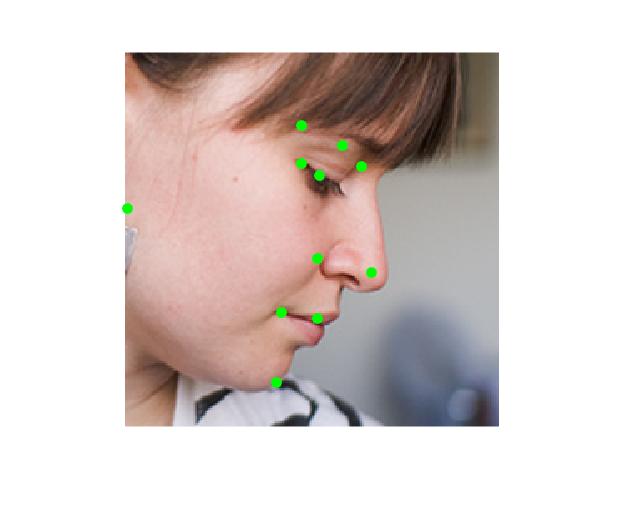}\includegraphics[width=3cm,height=3cm]{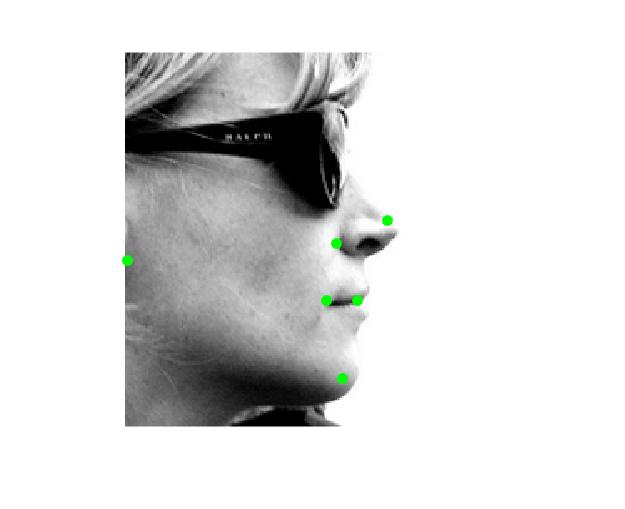}\includegraphics[width=3cm,height=3cm]{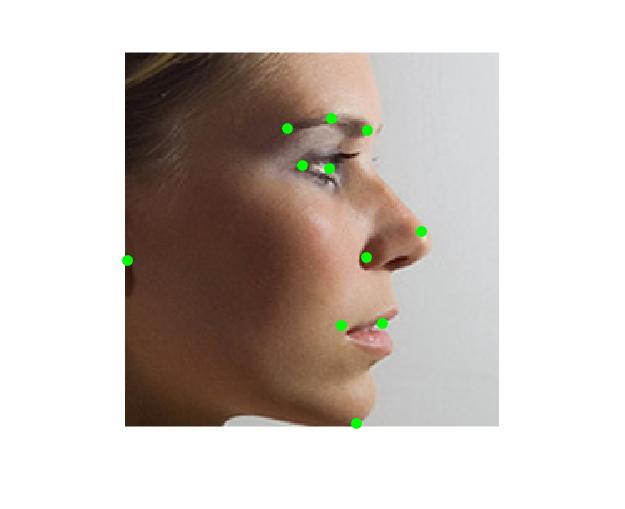}\includegraphics[width=3cm,height=3cm]{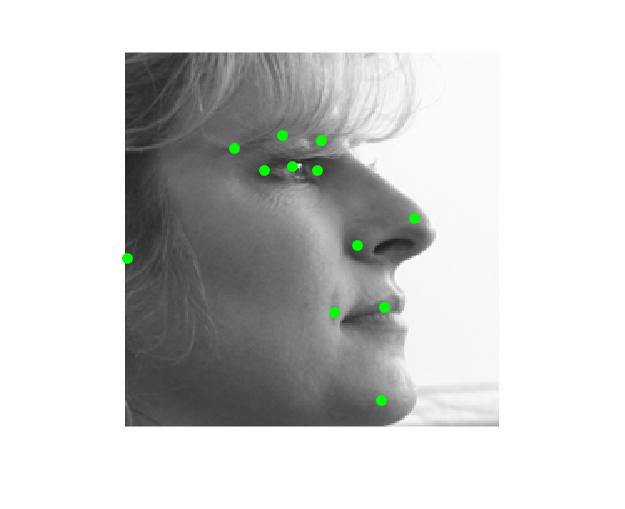}\includegraphics[width=3cm,height=3cm]{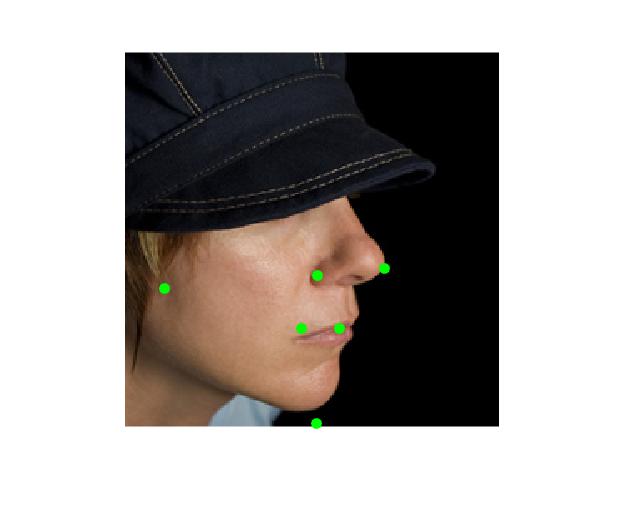}\includegraphics[width=3cm,height=3cm]{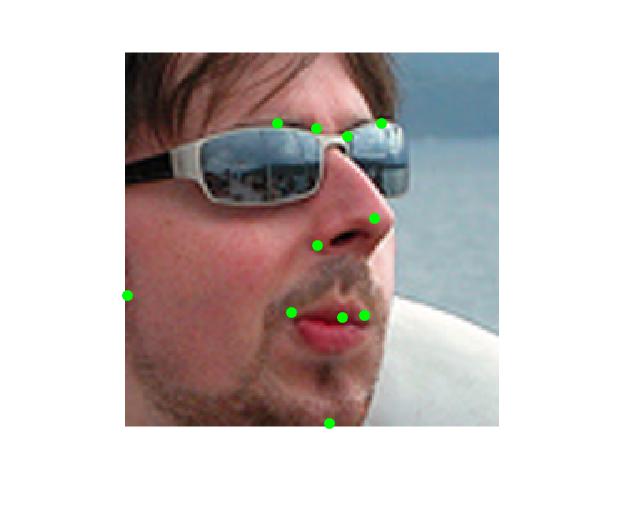}\\
\includegraphics[width=3cm,height=3cm]{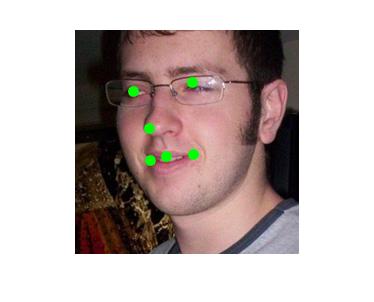}\includegraphics[width=3cm,height=3cm]{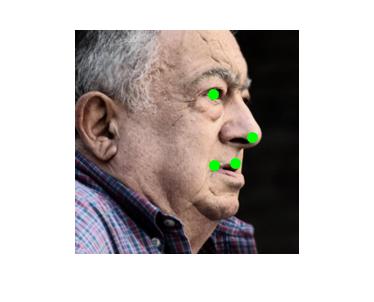}\includegraphics[width=3cm,height=3cm]{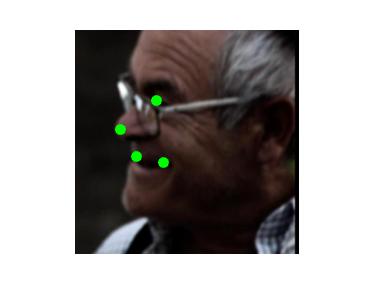}\includegraphics[width=3cm,height=3cm]{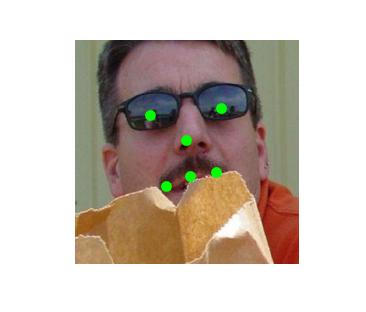}\includegraphics[width=3cm,height=3cm]{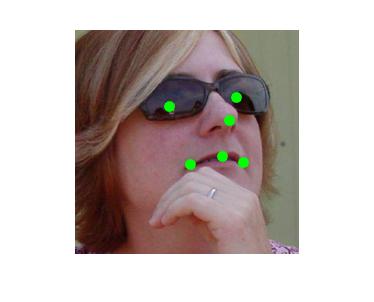}\includegraphics[width=3cm,height=3cm]{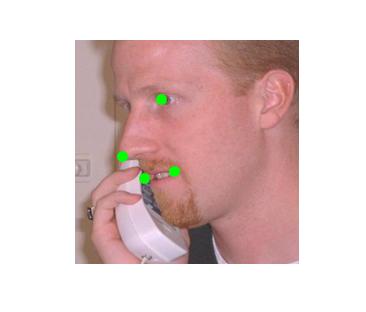}
\caption{Qualitative results generated from the proposed method. The green dots represent the predicted points. First row are the test samples from AFLW. Second row shows the samples from AFW dataset.}
\label{fig:qualitative}
\end{figure*}
\begin{figure}[htp!]
\centering
\includegraphics[height=5.5cm,width=8cm]{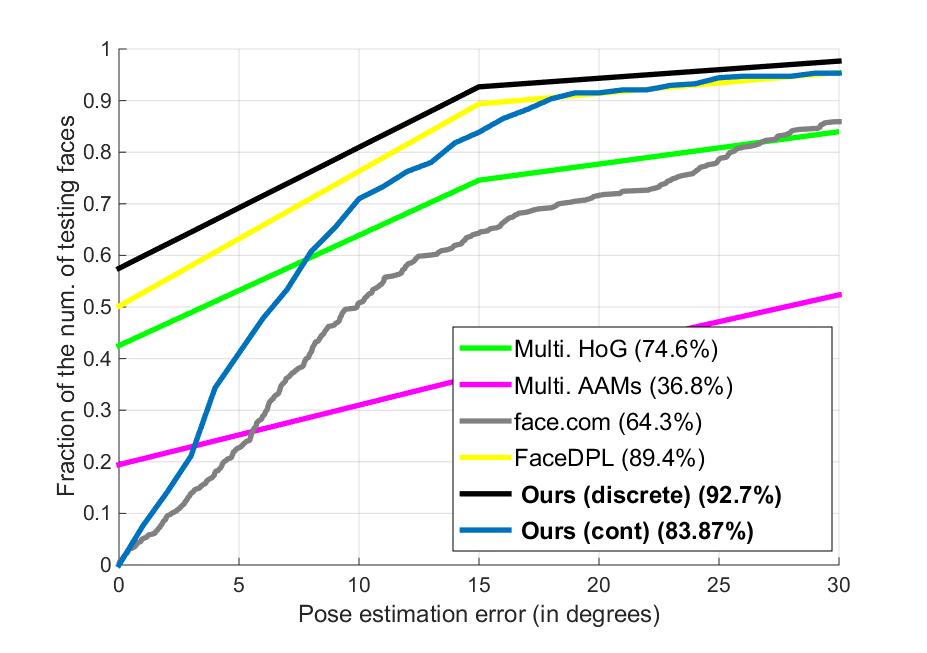}
\caption{Cumulative error distribution curves for pose estimation on AFW
dataset. The numbers in the legend are the percentage of faces that are
labeled within $\pm15\degree$ error tolerance}
\label{afw_pose}
\end{figure}

\textbf{Results:} Table \ref{aflw_table} compares the performance of proposed method over other existing methods. Table \ref{summary} summarizes the performance of the proposed method over different protocols. Figures \ref{aflw_res} and \ref{afw_res} shows the cumulative error distribution in predicting on the AFLW and AFW test sets. Figure \ref{afw_pose} shows the cumulative error distribution in pose estimation on AFW. Figure \ref{fig:qualitative} shows some qualitative results from AFLW and AFW testsets. In cases where a partial face is present inside the face bounding box, high probability on the heatmap was observed at keypoints of the partial face, in absence of message passing layers. This was effectively mitigated by the proposed message passing convolution layers. 

It is clear from the table that the proposed method outperforms other state-of-the art methods on AFLW dataset. It also outperforms all other methods except CCL\cite{Zhu_2016_CVPR} on the AFW dataset. \cite{Zhu_2016_CVPR} reannotates the AFLW dataset with $19$ points along with the invisible points leaving the ear points. The evaluation on AFW requires the coordinates of invisible points to be calculated as well. However, in our method invisible points are marked as background during training. Such invisible points are only constrained by the convolutional relationships models and constrained shape prediction. We conjecture that training on the reannotated data and using deeper models for regression, would make the prediction of occluded points more precise. 
\section{Conclusion}
In this work, we propose the idea of head pose conditioned modeling of correlations among different fiducial points for the task of face alignment through convolution-deconvolution networks. A feature level transfer of messages among keypoints delivers more detailed description about other keypoints than score maps. However, since for face images in extreme poses, information received from unstable keypoints is not meaningful, we propose conditioning the transfer of information based on head pose. A tree structure model is proposed for an effective message transfer. We show that without using cascade regression or 3D modeling approach, the proposed method not only performs comparable to those methods for the task of keypoint detection, but also for 3D head pose estimation. We tackle the problem of large number of parameters by using Residual Squeezenet networks. 

The proposed method provides a general framework that can be further applied to other localization specific tasks such as human pose estimation. In future, we plan to explore the proposed method for other tasks and for broader impact.
   

{\small
\bibliographystyle{ieee}
\bibliography{ref}
}

\end{document}